\documentclass[letterpaper, 10 pt, conference]{ieeeconf}  
\IEEEoverridecommandlockouts                              

\usepackage{graphics} 
\usepackage{epsfig} 

\usepackage{stfloats}
\usepackage{times} 
\usepackage{amsmath} 
\usepackage{amssymb}  

\usepackage{subcaption}
\usepackage{multirow}
\let\labelindent\relax
\usepackage{enumitem}

\usepackage{cite}
\usepackage{siunitx}
\usepackage{soul}
\usepackage{color}
\usepackage{xcolor}
\usepackage{mathrsfs}
\usepackage[ruled,linesnumbered]{algorithm2e}
\usepackage[font=scriptsize]{caption} 

\usepackage{acronym}

\soulregister\ref7
\soulregister\cite7 
\soulregister\eqref7
\soulregister\prettyref7

\usepackage{algpseudocode}
\usepackage{makecell}
\usepackage{booktabs}
\usepackage{threeparttable}
\usepackage{hyperref}
\hypersetup{hidelinks}
\usepackage{cleveref}
\usepackage{endnotes}

\bstctlcite{IEEEexample:BSTcontrol}

\usepackage{prettyref}
\newrefformat{fig}{Fig.~\ref{#1}}
\newrefformat{sec}{Sec.~\ref{#1}}
\newrefformat{tab}{Tab.~\ref{#1}}
\newrefformat{algo}{Algo.~\ref{#1}}
\newrefformat{eq}{(\ref{#1})}

\usepackage{color}

\title{\LARGE \bf
  Preference-Conditioned Multi-Objective RL for Integrated Command Tracking and Force Compliance in Humanoid Locomotion}

\author{Tingxuan Leng$^{1}$, Yushi Wang$^{1}$, Tinglong Zheng$^{2}$, Changsheng Luo$^{1}$, and Mingguo Zhao$^{1}$
\thanks{*This research was supported by STI 2030—Major Projects grant number 2021ZD0201402 and Beijing Natural Science Foundation(L243004)}
\thanks{$^{1}$Tsinghua University, Beijing 100084, China}%
\thanks{$^{2}$Beijing Jiaotong University, Beijing 100044, China}%
}

\begin{document}
\maketitle

\thispagestyle{empty}
\pagestyle{empty}

\begin{abstract}

Humanoid locomotion requires not only accurate command tracking for navigation but also compliant responses to external forces during human interaction. Despite significant progress, existing RL approaches mainly emphasize robustness, yielding policies that resist external forces but lack compliance-particularly challenging for inherently unstable humanoids. In this work, we address this by formulating humanoid locomotion as a multi-objective optimization problem that balances \textit{command tracking} and \textit{external force compliance}. We introduce a preference-conditioned multi-objective RL (MORL) framework that enables a single omnidirectional locomotion policy to trade off between command following and force compliance via a user-specified preference input. External forces are modeled via velocity-resistance factor for consistent reward design, and training leverages an encoder-decoder structure that infers task-relevant privileged features from deployable observations. We validate our approach in both simulation and real-world experiments on a humanoid robot. Experimental results in simulation and on hardware show that the framework trains stably and enables deployable preference-conditioned humanoid locomotion. Video can be found in the \href{https://lengtx20.github.io/morl-humanoid-locomotion/}{\textit{link}}.

\end{abstract}

\section{Introduction}

Humanoid robots are increasingly deployed in human-centered environments, making robust and interactive locomotion a central capability. Beyond robustness, locomotion must support natural physical interaction: humanoids should reliably follow commanded motions while remaining responsive to external guidance from humans. This capability, commonly referred to as \textit{force compliance}, is essential for practical deployment in scenarios involving physical guidance and close human–robot interaction. Without sufficient compliance, humanoids may exhibit overly rigid or potentially unsafe behaviors, such as resisting human intervention or responding with excessive forces.

Reinforcement learning with massively parallel simulation has enabled robust locomotion policies \cite{rudin2022learning, zhang2024resilient, hwangbo2019learning, gu2024humanoid, wang2025booster}. Recent work demonstrated walking on challenging terrains \cite{lee2020learning, kumar2021rma, kumar2022adapting, gu2024advancing}, blind navigation over stairs, push recovery \cite{melo2022learning, zhuang2024humanoid, xie2025humanoid}, and stylized walking from human references \cite{zhang2024whole, liao2025beyondmimic}, showcasing RL’s versatility for humanoid locomotion.

A key element of these approaches is the introduction of random force perturbations during training, which encourages the learned policy to remain stable under all conditions\cite{rudin2022learning, gu2024humanoid, wang2025booster}, but often biases policies toward resisting external forces, since stability under perturbations is explicitly rewarded. This training setting is essential for sim-to-real transfer, and enables reliable velocity-tracking with push recovery, while simultaneously hindering force-compliance behavior.

\begin{figure}
    \centering
    \includegraphics[width=1.0\linewidth]{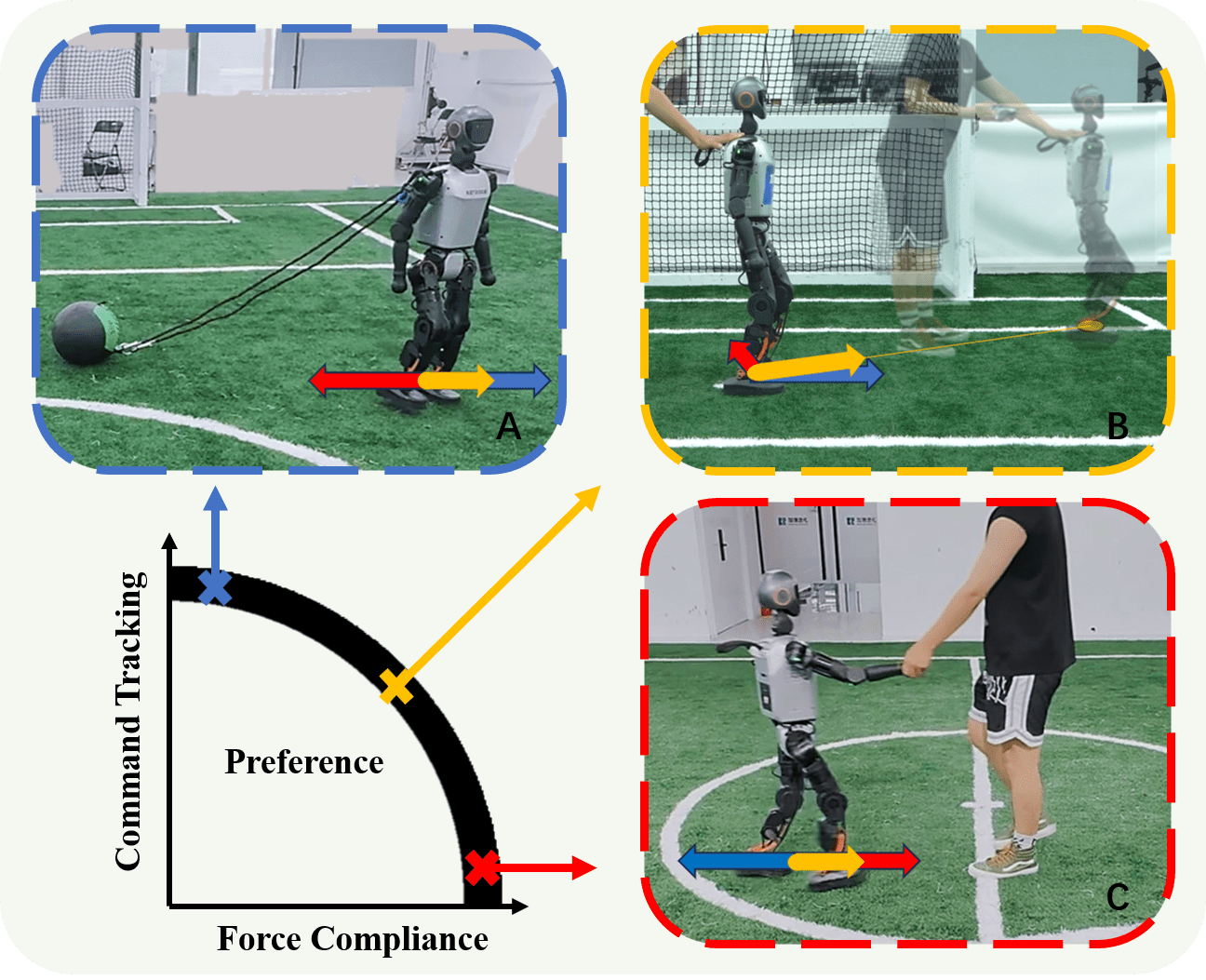}
    \caption{\textbf{Preference-conditioned locomotion:} A single policy realizes behaviors from command tracking to human-guided compliance by adjusting the preference. Arrows indicate velocity command (blue), external force (red), and resulting humanoid velocity (yellow).}
    \label{fig:title}
\end{figure}

\begin{figure*}
\vspace*{0.07in}
    \centering
    \includegraphics[width=\linewidth]{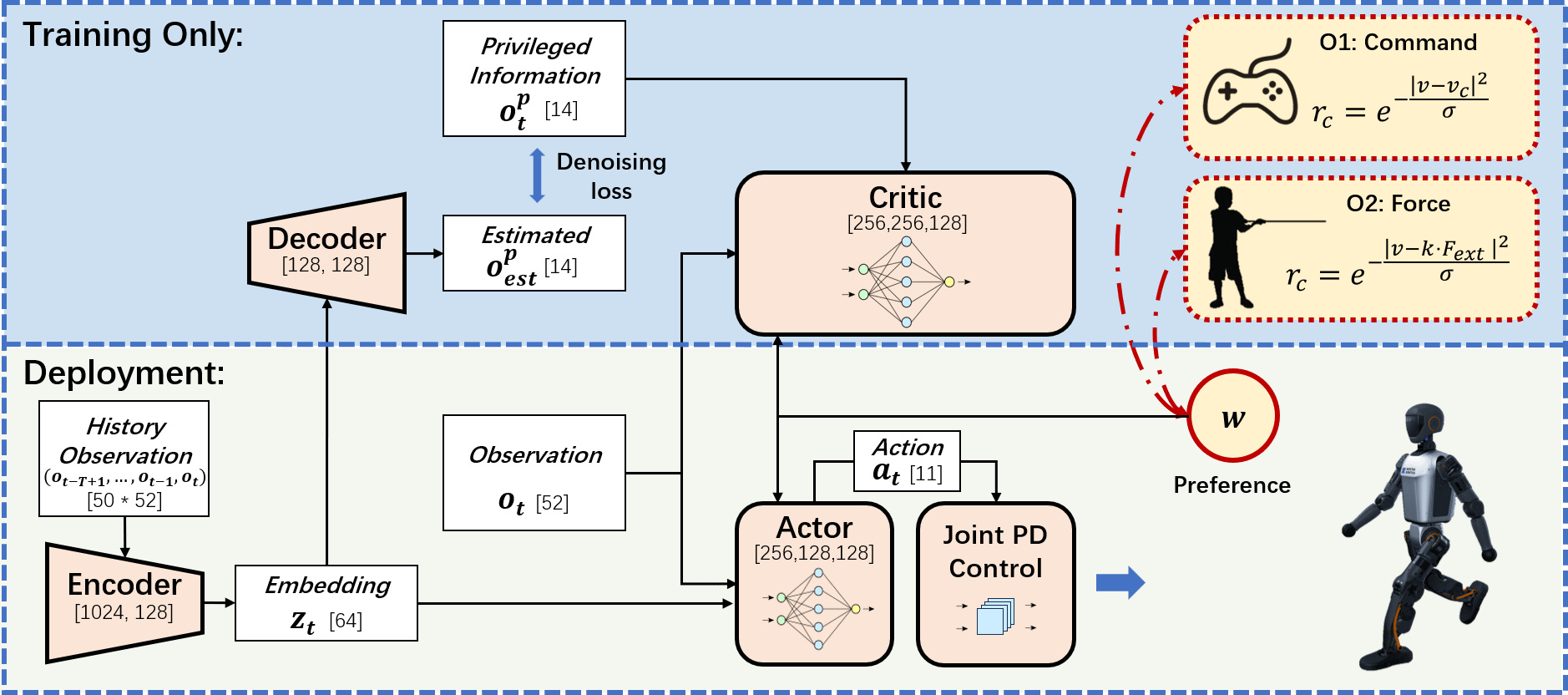}
    \caption{\textbf{Policy training framework with auxiliary reconstruction of privileged observations:} 
An asymmetric actor–critic architecture is extended with an encoder–decoder that reconstructs privileged observations, guiding the encoder to extract force- and torque-aware latent features. 
At deployment, only encoder and actor remain, enabling preference-conditioned control with onboard observations, latent embedding and preference vector.}
    \label{fig:framework}
\end{figure*}

Additionally, the pursuit of command tracking and force compliance is conflicting in nature: strong command following reduces compliance, while high compliance compromises responsiveness to current commands. Existing learning-based methods do not explicitly address this trade-off, and often compromise task-specific rewards to improve general performance \cite{hwangbo2019learning, margolis2023walk}, leaving a gap between robust command tracking and interactive, compliant walking.

To bridge this gap, we regard external force not simply as perturbation but as a balancing force in continuous compliant walking, and formulate humanoid locomotion with the trade-off as a multi-objective optimization problem. A preference-conditioned MORL framework is introduced with the conflicting objectives: (1) following velocity commands, and (2) complying with sustained external forces. By conditioning the locomotion policy on a preference vector, a single integrated policy can smoothly interpolate between rigid tracking and highly compliant walking. This design allows humanoids to achieve both capabilities without requiring additional training stages or complex hierarchical architecture.

We evaluate the proposed method in simulation and on the adult-size humanoid Booster T1. Results validate that our approach is able to produce deployable omnidirectional locomotion that adapts flexibly to different given preferences. Unlike conventional policies that focus solely on disturbance resistance, our method enables humanoids to operate under a broader range of interaction preferences in human-centered environments. The main contributions of this work are three-fold:

\begin{itemize}[topsep=1pt,itemsep=1pt,leftmargin=*]
\item We formulate humanoid locomotion as a \textbf{multi-objective problem}, where a \textbf{velocity–resistance model} provides a unified representation of commands and external forces, enabling consistent and comparable reward design.
\item We propose a \textbf{preference-conditioned MORL framework} that trains a \textbf{single policy} to cover a continuous spectrum of tracking–compliance trade-offs, without hierarchical controllers or staged training.
\item We validate the proposed approach in both simulation and real-world experiments on a humanoid robot, demonstrating deployable locomotion behaviors across different user-specified interaction preferences.

\end{itemize}

\section{Related Work}
\subsection{RL-based Humanoid Locomotion}

Reinforcement Learning has become a prominent approach for humanoid locomotion, with policies trained in large-scale parallel simulation and deployed zero-shot on hardware. Open-source frameworks \cite{rudin2022learning, gu2024humanoid, wang2025booster} provide standardized environments and scalable pipelines, accelerating progress on robust and deployable locomotion policies for quadrupeds and humanoids.

A key paradigm in recent works is the \textit{asymmetric actor–critic} architecture, where the critic has access to full states while the actor is constrained to partial observations. Teacher–student schemes further leverage historical observations to help the student policy infer task-relevant features. For example, RMA \cite{kumar2021rma} adopts a two-stage pipeline: a teacher policy is trained in the first stage with privileged information only available in simulation, and a student policy is distilled to rely only on deployable observations and latent embeddings. A-RMA \cite{kumar2022adapting} extends this to a three-stage procedure with an additional policy refinement process to adapt to a bipedal robot. More recently, \cite{gu2024advancing} streamlined this paradigm into a single-stage framework, jointly training encoder and policy to enable zero-shot humanoid deployment. We build upon these advances and adopt Booster Gym \cite{wang2025booster} as the fundamental codebase and baseline environment.

\subsection{Multi-Objective RL for Robotics}

MORL has been widely applied in robotics, where objectives naturally conflict \cite{shu2024learning}. In animation and motion generation, it has been used to balance fidelity, style, and user control, as in AMOR \cite{alegre2025amor} and \cite{shrestha2024generating}, demonstrating strong performance across both robotics and animation domains. Furthermore, studies of biological and robotic gaits highlight inherent trade-offs among speed, energy, and stability \cite{rockenfeller2024coordinating}, which further motivates the use of MORL in locomotion.  

Building on this perspective, recent work has applied MORL to legged robots, coordinating objectives such as velocity tracking, torque cost, robustness, and whole-body control \cite{kim2025stage, robert2024multi, tran2023two}. While these approaches show promising results, the objectives are typically \textit{indirectly} conflicting—such as speed versus energy efficiency or fidelity versus robustness—where improvements often emerge from smooth trade-offs rather than explicit opposition.

Instead, our work focuses on balancing velocity command tracking and force compliance within a single humanoid locomotion policy, which poses practical challenges for deployable control.

\subsection{Locomotion with Force Adaptation}
Learning-based approaches have begun to address locomotion under external forces. Advances in RL-based method has enabled legged robots to remain robust under perturbations and heavy loads \cite{rudin2022learning, kumar2021rma} and exert force needed for loco-manipulation tasks\cite{portela2024learning, fu2023deep}. However, force compliance is overlooked: while training policies to resist external forces, it makes compliant human guidance difficult and often produces unnatural reactions.

To encourage force-adaptive behaviors for legged robots, several works propose dedicated RL-based methods, while primarily focusing on quadruped locomotion or loco-manipulation tasks. Hartmann et al.\cite{hartmann2024deep} introduced compliant quadruped control through a recovery stage during training. HAC-LOCO \cite{zhou2025hac} adopts a hierarchical design where a high-level residual module modulates a robust low-level policy. FACET \cite{xu2025facet} instead imitates a spring–impedance reference through a two-stage teacher–student pipeline with adaptive parameters. FALCON \cite{zhang2025falcon} addresses loco-manipulation with a dual-agent design and torque-limit-aware 3D end-effector force curricula, while the lower agent focusing on the supporting manipulation goals. While these methods demonstrate effective adaptation in their respective domains, their assumptions and architectures are not directly tailored to humanoid omnidirectional locomotion under sustained human-applied forces. In contrast, our work addresses humanoid locomotion exclusively, aiming to unify velocity command tracking and external force compliance in a single policy. Our method demonstrates preference-conditioned force adaptability in humanoid omnidirectional locomotion within a unified learning-based framework.

\section{Method}

In this section, we present our method for learning a humanoid locomotion policy that balances command tracking and force compliance. We first align the velocity and force from the perspective of a consistent velocity-resistance model (\Cref{method:vel-model}). Then, we formulate the problem as a MORL task with preference conditioning (\Cref{method:morl}). Building on these, we establish our training framework to obtain adaptive and deployable policy (\Cref{method:frame}).

\subsection{Velocity-Resistance Modeling of External Force}
\label{method:vel-model}

We formulate humanoid locomotion as balancing velocity tracking and force compliance in the horizontal plane, while vertical dynamics, which is commonly related with complex terrains, is beyond the scope of this work. Both linear and angular velocities are considered, corresponding respectively to external forces and torques. In practice, we envision the robot to respond in a steady and continuous manner when guided by external forces, e.g., maintaining gentle contact with a human hand during interaction.

To balance command tracking and force compliance within a single locomotion policy, velocity commands and external forces must be expressed in a comparable form. However, they inherently lie in different physical spaces. To unify them, we adopt a velocity–resistance perspective that maps sustained external forces into equivalent velocities, allowing both objectives to be represented consistently in the reward function.

In many physical systems, resistive forces scale linearly with velocity, as in viscous damping:
\begin{equation}
    F_{\text{res}} = -B \cdot v,
\end{equation}
where \(B\) is an effective damping coefficient. At steady state, external and resistive forces balance, yielding
\begin{equation}
    F_{\text{ext}} + F_{\text{res}} \approx 0 \quad\Longrightarrow\quad v \approx B^{-1} \cdot F_{\text{ext}}.
\end{equation}

This formulation is physically suitable for humanoid locomotion: (i) human-applied forces are typically slow and low-frequency, making the steady-state approximation reasonable; (ii) it induces intuitive compliance—robots move only while being pulled, and stop naturally once the force is released; and (iii) it avoids oscillatory or unstable behaviors that can arise from high-order dynamics.

For training, we adopt a simplified mapping
\begin{equation}
    v_{\text{ext}} = k \cdot F_{\text{ext}},
\end{equation}

In practice, \(k\) is selected such that the resulting equivalent velocity $v_{ext}$ falls within the typical range of commanded base velocities during training. We found the policy behavior to be qualitatively consistent across a moderate range of k values, and therefore fix k for all experiments.

This approximation is not intended to replace full impedance control for high-bandwidth interaction, but rather provides a stable, interpretable, and physically grounded mechanism that makes external forces directly comparable to velocity commands during locomotion training with RL.

\subsection{Multi-Objective Reinforcement Learning Formulation}
\label{method:morl}
We formulate humanoid locomotion as a multi-objective reinforcement learning problem. 

The standard locomotion problem is modeled as a Partially
 Observable Markov Decision Process (POMDP) $(\mathcal{S}, \mathcal{O}, \mathcal{A}, p,r,\gamma)$, where $\mathcal{S}, \mathcal{O}, \mathcal{A}$ denote the state, observation, and action spaces, $p(s'|s,a)$ the dynamics, $r$ the reward, and $\gamma$ the discount. Policy is represented by $\pi(a|o)$. In MORL, the reward is extended to a vector
\begin{equation}
    \mathbf{r}(s,a) = [r_1(s,a), \dots, r_n(s,a)],
\end{equation}
capturing multiple, possibly conflicting objectives. The goal is to approximate a family of balanced trade-offs. To this end, we adopt a preference-conditioned policy $\pi(a|o, \mathbf{w})$ with weight vector $\mathbf{w}=[w_1,\dots,w_n], \text{s.t. }\sum_i w_i=1$. The policy maximizes the expected return:
\begin{equation}
    J(\pi; \mathbf{w}) = \mathbb{E}_{\tau \sim p_{\pi}} \Big[\sum_{t=0}^\infty \gamma^t \, \mathbf{r}(s_t,a_t)\cdot \mathbf{w}\Big].
\end{equation}

Sampling diverse $\mathbf{w}$ during training enables the policy to cover a family of solutions.

For humanoid locomotion, we consider three objectives: \textbf{command tracking} ($r_c$), \textbf{force compliance} ($r_f$), and \textbf{regularization} ($r_r$):
\begin{equation}
    r_c = \exp\Big(-\tfrac{\|v - v_{c}\|^2}{\sigma}\Big),
\end{equation}
\begin{equation}
    r_f = \exp\Big(-\tfrac{\|v - k\cdot F_{\text{ext}}\|^2}{\sigma}\Big),
\end{equation}
where $\sigma$ is a hyperparameter. $r_c$ measures the accuracy of command tracking, and $r_f$ quantifies compliance to external forces in a comparable approach. The regularization term $r_r$ is the sum of all additional rewards (e.g., base height, energy cost and stability penalties). For our task, $w_c$ and $w_f$ are constrained such that $w_c + w_f = 2$.

\subsection{Policy Training with Privileged Reconstruction}
\label{method:frame}
Learning force compliance is challenging since external forces cannot be directly measured without tactile sensors. Inspired by \cite{gu2024advancing}, we design a one-stage privileged reconstruction framework, where the encoder infers force-related features from historical observations (see \Cref{fig:framework})

\subsubsection{Observation and Action Space}
The actor receives deployable observations $o$ (proprioception, velocity commands, gait features, etc.), while the critic additionally accesses privileged information $o^p$ (e.g., external forces, linear velocities), available only in simulation. The action is the desired joint position $a=q_{des}$. A detailed specification of observations is shown in \Cref{tab:notations1}.

\vspace{0pt}
\begin{table}[!ht]
\vspace{0pt}
\caption{Observation Space}
\label{tab:notations1}
\renewcommand{\arraystretch}{1.3}
\begin{center}
\begin{tabular}{l l l l}
\hline
\textbf{Components} & \textbf{Dim} & \textbf{Critic} & \textbf{Actor}  \\
\hline
Preference Vector & 3 & \checkmark & \checkmark \\
Velocity Commands & 3 & \checkmark & \checkmark \\
Projected Gravity & 3 & \checkmark & \checkmark \\
Angular Velocity & 3 & \checkmark & \checkmark \\
Gait Signal & 2 & \checkmark & \checkmark \\
Gait Frequency & 1 & \checkmark & \checkmark \\
Joint Position & 12 & \checkmark & \checkmark \\
Joint Velocity & 12 & \checkmark & \checkmark \\
Last Action & 12 & \checkmark & \checkmark \\
\hline
Body Center of Mass & 3 & \checkmark &  \\
Body Mass & 1 & \checkmark &  \\
Linear Velocity & 3 & \checkmark &  \\
Body Height & 1 & \checkmark &  \\
Pushing Force & 3 & \checkmark &  \\
Pushing Torque & 3 & \checkmark &  \\
\hline
\end{tabular}
\end{center}
\vspace{-6pt}
\end{table}

\subsubsection{Network Architecture and Training Pipeline}

We adopt an asymmetric actor–critic structure with an encoder–decoder module. The encoder maps the stacked historical observations $o_H = (o_{t-T+1}, ..., o_{t-1}, o_{t})$ into a latent embedding $z_t$, and the decoder reconstructs privileged states $o^p_t$, encouraging $z_t$ to encode task-relevant features. The actor is conditioned on $(o_t, z_t, w_t)$ and outputs joint commands tracked by a low-level joint PD controller, while the critic has access to both observations and privileged observations $(o_t,o_t^p)$ without requiring the embedding.

During training, the actor and critic are updated using PPO\cite{schulman2017proximal}, with the MORL rewards defined in \Cref{method:morl}. For additional implementation details of the rewards, please refer to \cite{wang2025booster}. To obtain a deployable policy, we employ a curriculum learning strategy together with domain randomization (DR). Training begins on flat terrain without velocity perturbations (sudden velocity impulses), which are gradually increased as training progresses. The environment is then switched to uneven terrain after half of the training. External forces are randomly applied during training, resampled every 30~s from $U(-50,50)$ and later narrowed to $U(-20,20)$ to increase sensitivity to subtle interactions. The preference weight $w_c$ is resampled each episode from $U(0,2)$ with $w_f = 2.0 - w_c$,  while $w_r$ is randomly sampled from $U(1,2)$ to improve robustness.

The encoder–decoder is optimized jointly with reconstruction losses:
 \begin{equation}
     L_{rec} = \| \hat{o^p} - o^p\|^2
 \end{equation}
 \begin{equation}
      L_{rec,force} = \| \hat{F_{\text{ext}}} - F_{\text{ext}}\|^2
 \end{equation}
where $\hat{o^p}$ is the estimated privileged observation. The second term is introduced to emphasize force-related features.

At deployment, only the encoder and actor are used, producing a policy that adapts online according to the preference vector. For evaluation, the regularization weight is fixed to $w_r = 1$ for consistency and better robustness.

\section{Simulation Experiments and Analysis}

The policy is trained in Isaac Gym \cite{makoviychuk2021isaac} with 4096 parallel environments. Each episode lasts 20 seconds with conditional early termination, and training is run for 20,000 epochs. All experiments are performed on the Booster T1 humanoid model at a 50~Hz control frequency, and are further validated via sim-to-sim transfer to MuJoCo~\cite{todorov2012mujoco}.

We evaluate our approach against a humanoid locomotion baseline built on Booster Gym \cite{wang2025booster}, which provides a standardized end-to-end RL pipeline and serves as a stable baseline for sim-to-real humanoid locomotion. The baseline policy adopts the same network structure and training pipeline, including domain randomization and external perturbations, but is trained without an explicit force-compliance objective. This comparison isolates the effect of explicitly modeling force compliance through preference-conditioned multi-objective training. 

In the following sections, we first examine whether our preference-conditioned MORL policy (MORL policy in short) can learn a set of behaviors that captures the trade-off between command tracking and force compliance (Sec.~\ref{sec:trade-off}). We then evaluate its adaptability when preference weights are switched online (Sec.~\ref{sec:online}), compare its training behavior against a single-objective RL pipeline (Sec.~\ref{sec:ablation}), and finally assess its robustness under unexpected perturbations (Sec.~\ref{sec:simrobust}).

\subsection{Trade-off Performance across Different Preferences}
\label{sec:trade-off}
To demonstrate its trade-off behavior and adaptability across multiple objectives, we evaluate the policy under command weights from $0.0$ to $2.0$ with interval $0.1$ and corresponding $w_f = 2-w_c$.

First, the robot operates in the opposite setting: it is commanded to walk forward with a constant velocity while subjected to three levels of backward force. The average forward velocity $\bar v_x$ is recorded for each trial. Similarly, angular velocity tracking is tested with constant yaw commands under three levels of external torque. For visualization, rewards for each objectives are expressed as mean squared errors (MSE):
\begin{equation}
    r_c = \mathbf{MSE}(v_x, v_{c,x})
\end{equation}
\begin{equation}
    r_f = \mathbf{MSE}(v_x, k\cdot F_{\text{ext}})
\end{equation}

\begin{figure}
    \vspace*{0.07in}
    \centering
    \includegraphics[width=1.0\linewidth]{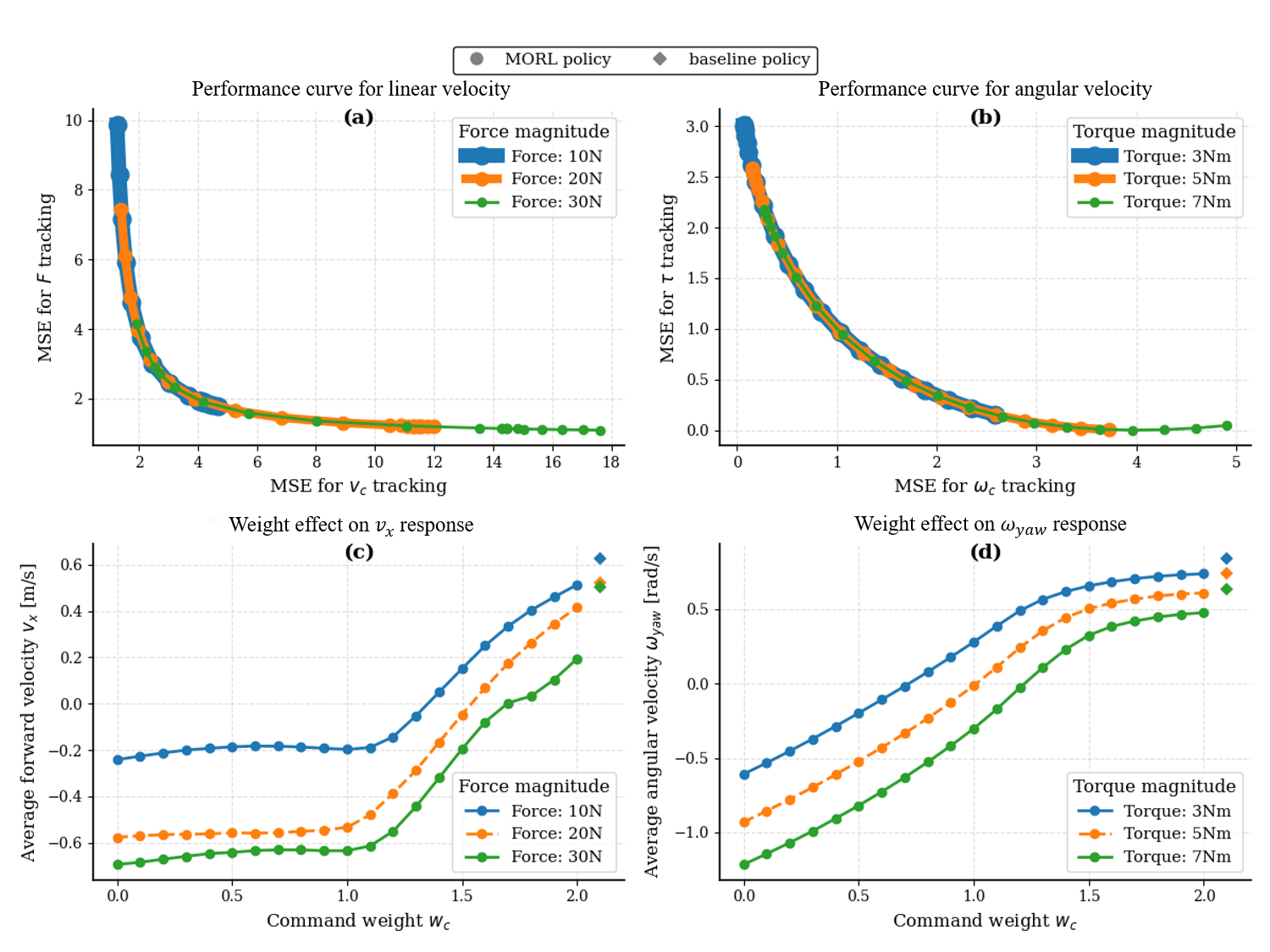}
    \caption{\textbf{Effect of reward weight $\mathbf{w}$ in the opposite setting:} Each trial lasts 5~s, and the baseline policy result is plotted on the rightmost side for comparison.
    (a)(b) Performance curve visualization across different preference weights for linear and angular velocity.
    (c) average forward velocity $\bar{v}_x$ with $v_{c,x}=1.0 \ \mathrm{m/s}$ under three levels of backward force (10~N, 20~N, 30~N). 
    (d) average yaw angular velocity $\omega_{yaw}$ with $\omega_{c}=1.0 \ \mathrm{rad/s}$under three levels of applied torque (3~Nm, 5~Nm, 7~Nm).
    }
    \label{fig:expmorl}
\end{figure}

For visualization clarity, the reported MSE values are computed over time-averaged steady-state windows after initial transients, and smoothed using an exponential moving average.
The results in \Cref{fig:expmorl} show trade-off curves (upper panels) and velocity responses (lower panels). From the results, we observe that many solutions exhibit clear trade-offs, where improving one objective typically degrades the other. This suggests that preference-conditioned training produces a consistent empirical trade-off curve across different preference settings. The curve exhibits an approximately monotonically decreasing trend, consistent with the practical tension between command tracking and force compliance in this setting. Moreover, the smoothness of the curve suggests that our policy provides a continuous and feasible solution space, accommodating arbitrary user-specified preferences. 

\begin{figure}
    \vspace{-6pt}
    \centering
    \includegraphics[width=1.0\linewidth]{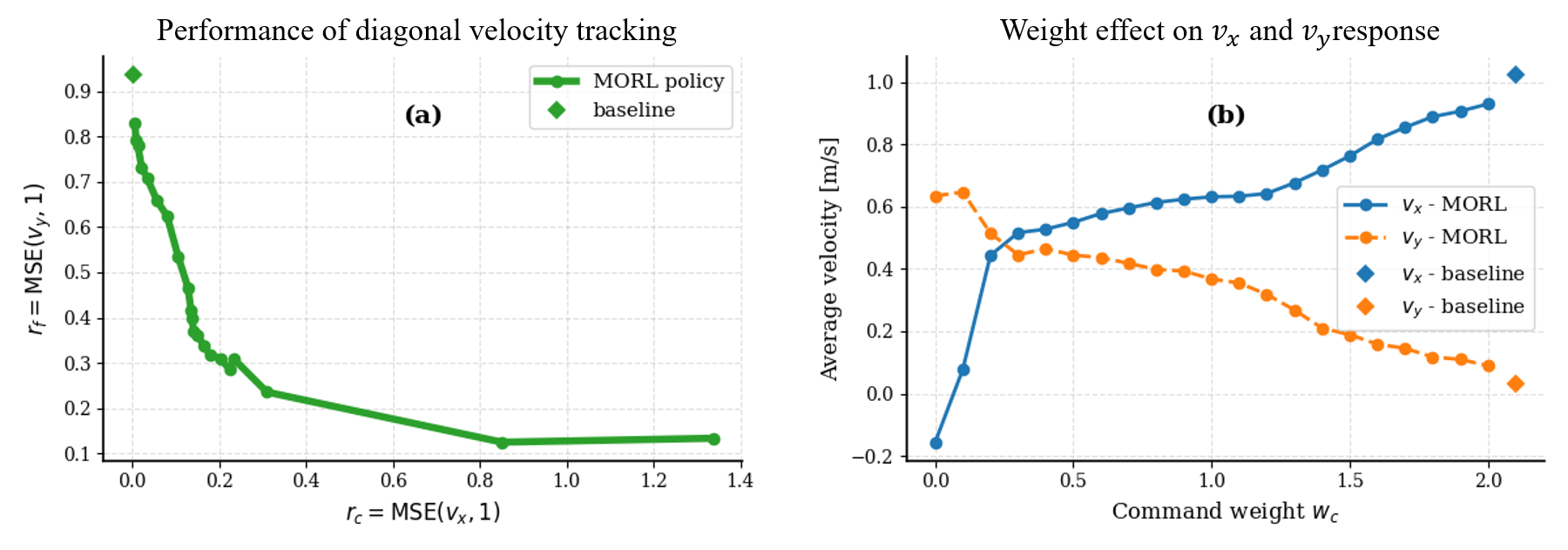}
    \caption{\textbf{Effect of reward weight $\mathbf{w}$ in the orthogonal setting:} Each trial lasts 5~s, and the baseline policy result is plotted for comparison. 
    (a) Trade-off curve visualization across different preference weights for linear velocity.
    (b) velocity $v_x$ and $v_y$ with $v_{c,x}=1.0 \ \mathrm{m/s}$ under a leftward force (30~N). }
    \label{fig:exp22morl}
\end{figure}

We further test an orthogonal setting where forward velocity is commanded while lateral forces are applied. As shown in \Cref{fig:exp22morl}, the policy produces diagonal walking by combining both objectives, whereas the baseline fails to adapt and mainly tracking command. 

During all experiments, the MORL policy succeeds without any unstable behaviors or falls, validating the reliability of our training setup. The policy demonstrates a smooth trade-off between tracking and compliance across preference weights. When $w_c = 2.0$ (and $w_f = 0.0$), the MORL policy achieves comparable tracking accuracy to the baseline. For smaller $w_c$, it exhibits significantly stronger compliance to external forces, which the baseline lacks due to its force-unaware training. These results confirm the effectiveness of the proposed method in integrating competing objectives and highlight its versatility in adapting to diverse interaction demands.

\subsection{Adaptation to Online Preference Switching}
\label{sec:online}
Furthermore, the MORL policy supports online switching of behaviors when preference weights are changed during deployment. The humanoid is tested under two separate conditions: either commanded with a constant forward velocity $v_{c,x}$ or subjected to a constant external force $F_{\text{ext}}$, with the forward velocity reward weight $w_c$ switched twice during each trial. Similarly, angular velocity and torque tracking are evaluated under analogous separate conditions.

\begin{figure}
    \vspace*{0.07in}
    \centering
    \includegraphics[width=1.0 \linewidth]{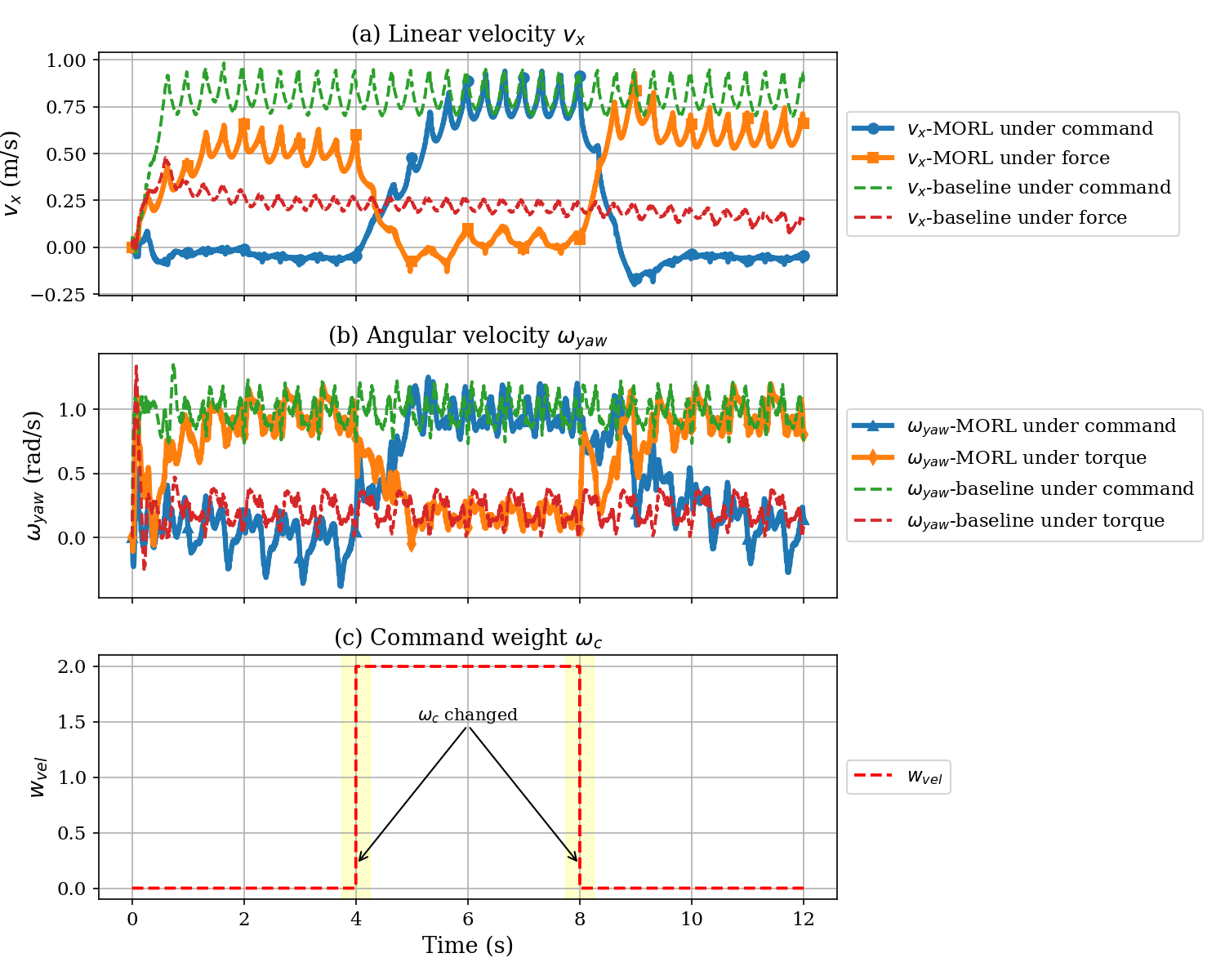}
    \caption{\textbf{Online switching of command weight:} The trajectory lasts for 12~seconds with the command weight changed every 4~seconds. The robot is applied constant $v_{c,x} = 1.0 \mathrm{m/s}, F_{\text{ext}}=20\mathrm{N}, \omega_c = 1.0 \mathrm{rad/s}, \tau=5\mathrm{N \cdot m}$ separately. The policy's behavior changes corresponding to the command weight.}
    \label{fig:expswap}
\end{figure}

As shown in \Cref{fig:expswap}, the policy transitions smoothly between behaviors without instability or sudden performance degradation, achieving performance comparable to the baseline on single-objective tracking. In contrast, the baseline policy supports only a single motion mode. These results highlight that the MORL framework not only provides a continuous spectrum of trade-offs but also enables seamless online preference adaptation, a practical capability for interactive deployment where task priorities may shift over time.

\subsection{Ablation Study: MORL vs. Single-Objective RL}
\label{sec:ablation}
\begin{figure*}
    \vspace*{0.07in}
    \centering
    \includegraphics[width=\linewidth]{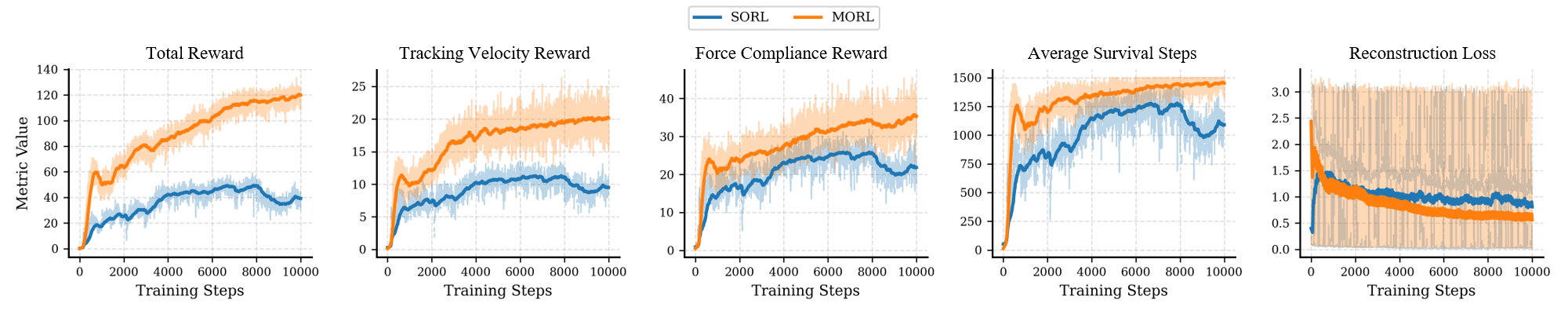}
    \caption{Ablation study comparing MORL and SORL policies across five metrics: (1) total reward, (2) tracking velocity reward, (3) force compliance reward, (4) average survival steps, and (5) auxiliary reconstruction loss. Curves are smoothed using Exponential Moving Average ($\alpha=0.01$) while having the original values faded in the background.}
    \label{fig:ablation}
\end{figure*}

To further validate the effectiveness of our approach, we conduct an ablation study against a standard single-objective RL (SORL) pipeline trying to capture multiple objectives. In the single-objective setting, the trade-off is modeled as a scalar sum of rewards for conflicting objectives, with fixed weights $w_c = w_f = w_r = 1$. Both MORL and SORL use identical network architectures, optimizer settings, reward functions and amount of training steps (10,000 epochs); only the preference sampling differ. The comparison of the training curves is reported in \Cref{fig:ablation}.

The MORL policy exhibits more stable training behavior than SORL across the reported metrics, while the auxiliary reconstruction loss also decreases smoothly. We observe that the MORL policy achieves stable training across a range of preference settings. While the reward curves are not directly comparable in magnitude due to different objective weightings, the results indicate that preference conditioning does not hinder convergence and enables a single policy to cover multiple trade-off behaviors. This training behavior likely stems from the inherent conflict between objectives. After training, the MORL agent develops adaptive locomotion skills. In contrast, the SORL policy exhibits limited robustness when trained with a fixed combined objective, highlighting the challenge of balancing conflicting objectives within a single scalar reward. These results suggest that explicitly modeling conflicting objectives with MORL provides a more stable and flexible training framework, particularly when a single fixed reward weighting may not be well suited for balancing competing behaviors.

\subsection{Robustness under Instant Perturbation}
\label{sec:simrobust}
The test duration for this experiment is 20 seconds. The perturbation is modeled as an external force with fixed magnitude but random orientation. At every 5-second interval, a 1-second force impulse is applied to the robot’s base link. We evaluate four different policy settings: (1) baseline policy, (2) MORL-c policy with $w_c, w_f = [2,0]$, (3) MORL-m policy with $w_c, w_f = [1,1]$, and (4) MORL-f policy with $w_c, w_f = [0,2]$. Three levels of force magnitude are tested: $F_{\text{ext}} = 30\text{N}, 40\text{N}, 50\text{N}$. Each experiment is repeated 20 times for statistical reliability.

Two metrics are used to evaluate policy performance:
\begin{itemize}
    \item \textbf{Success Rate:} The proportion of trials in which the humanoid doesn't fall during the testing time.
    \item \textbf{Peak Torque:} The average of the maximum root-mean-square (RMS) torque across all joints for each successful trial. For each trial, the RMS torque is computed over time for each joint, then the maximum RMS value among all joints is selected and are averaged over all successful trials. Only successful trials are considered to avoid spurious values from post-fall twitching.
\end{itemize}

\vspace{3pt}
\begin{table}[h!]
\centering
\scriptsize
\caption{Results of External Perturbation Experiments}
\label{tab:successrate}
\renewcommand{\arraystretch}{1.3}
\begin{tabular}{l | l l l l l}
\hline
\textbf{Force} & \textbf{Policy} & baseline & MORL-c & MORL-m & MORL-f \\
\hline
\multirow{2}{*}{30N} & Success Rate & 100\% & 100\% & 100\% & 100\% \\
                     & Peak Torque & 15.542 & 14.677 & 15.249 & 14.457 \\
                     & Std      & ±0.214  & ±0.146  & ±0.184  & ±0.174 \\
\hline
\multirow{2}{*}{40N} & Success Rate & 45\% & 90\% & 85\% & 100\% \\
                     & Peak Torque & 15.282 & 14.864 & 15.249 & 14.635 \\
                     & Std      & ±0.165  & ±0.317  & ±0.237  & ±0.394 \\
\hline
\multirow{2}{*}{50N} & Success Rate & 0\% & 35\% & 50\% & 50\% \\
                     & Peak Torque & -- & 14.758 & 15.229 & 14.589 \\
                     & Std      & -- & ±0.388 & ±0.201  & ±0.408 \\
\hline
\end{tabular}
\vspace{3pt}
\end{table}

The results in Table II indicate that while all policies perform equally well under mild perturbations (30N), MORL-based policies exhibit greater robustness at higher force magnitudes. In particular, MORL-f consistently achieves the highest success rates without increasing joint torque. 
We note that MORL-m does not minimize peak torque by design, as it balances command tracking and compliance simultaneously. In contrast, MORL-c prioritizes compliance and therefore exhibits lower peak torque under strong perturbations.
Even under 50N perturbations, MORL-m and MORL-f maintain a 50\% success rate, whereas the baseline fails completely. Moreover, in most cases, greater policy compliance corresponds to lower peak joint torque. These findings highlight the robustness advantage of MORL agents in terms of both stability and compliance.

\section{Real-World Experiments}

We validate the proposed MORL-based locomotion policy on the Booster T1 humanoid robot through a series of hardware experiments. These experiments are designed to demonstrate two key aspects: (1) adaptability to different objective preferences, and (2) quantitative force-compliance performance.

\subsection{Adaptability to Different Preferences}
To evaluate adaptability, we conduct experiments under three preference settings identical to those in \Cref{sec:simrobust}, with the robot guided either by joystick velocity commands or by sustained human-applied guiding forces.

Across these settings, the robot exhibits qualitatively distinct locomotion behaviors as the preference weights vary. As the force-compliance weight $w_f$ increases, the robot responds more readily to external forces, yielding to human guidance with reduced resistance. Conversely, higher command-tracking weights $w_c$ result in faster and more precise execution of joystick commands, accompanied by increased resistance to external forces. In contrast, the baseline policy lacks this flexibility, as it is optimized solely for velocity tracking and becomes unstable under sustained human-applied forces.

These observations demonstrate that preference conditioning enables flexible adjustment between command tracking and force-compliance behaviors within a single deployed policy. Representative demonstrations of these behaviors are provided in the supplementary video for visualization.

\subsection{Cross-Directional Walking}
Beyond balancing between tracking and compliance, we further evaluate whether the policy can integrate multiple objectives simultaneously. Drawing from observations in earlier experiments, we choose preference weights $w_c, w_f, w_r = [1.5, 0.5, 2.0]$, which provide a practical balance between command tracking and force compliance. With this configuration, the robot is commanded to walk forward or laterally while being subjected to orthogonal external forces, as illustrated in \Cref{fig:title}(b). 

Under these settings, the robot consistently produces diagonal walking, effectively combining the commanded velocity and the compliant response to external forces. This emergent behavior highlights the MORL policy’s ability not only to trade off between objectives, but also to synthesize them into coherent locomotion strategies in real time.

\subsection{Force-Compliance Locomotion}
\begin{figure}
    \vspace*{0.07in}
    \centering
    \includegraphics[width=1.0\linewidth]{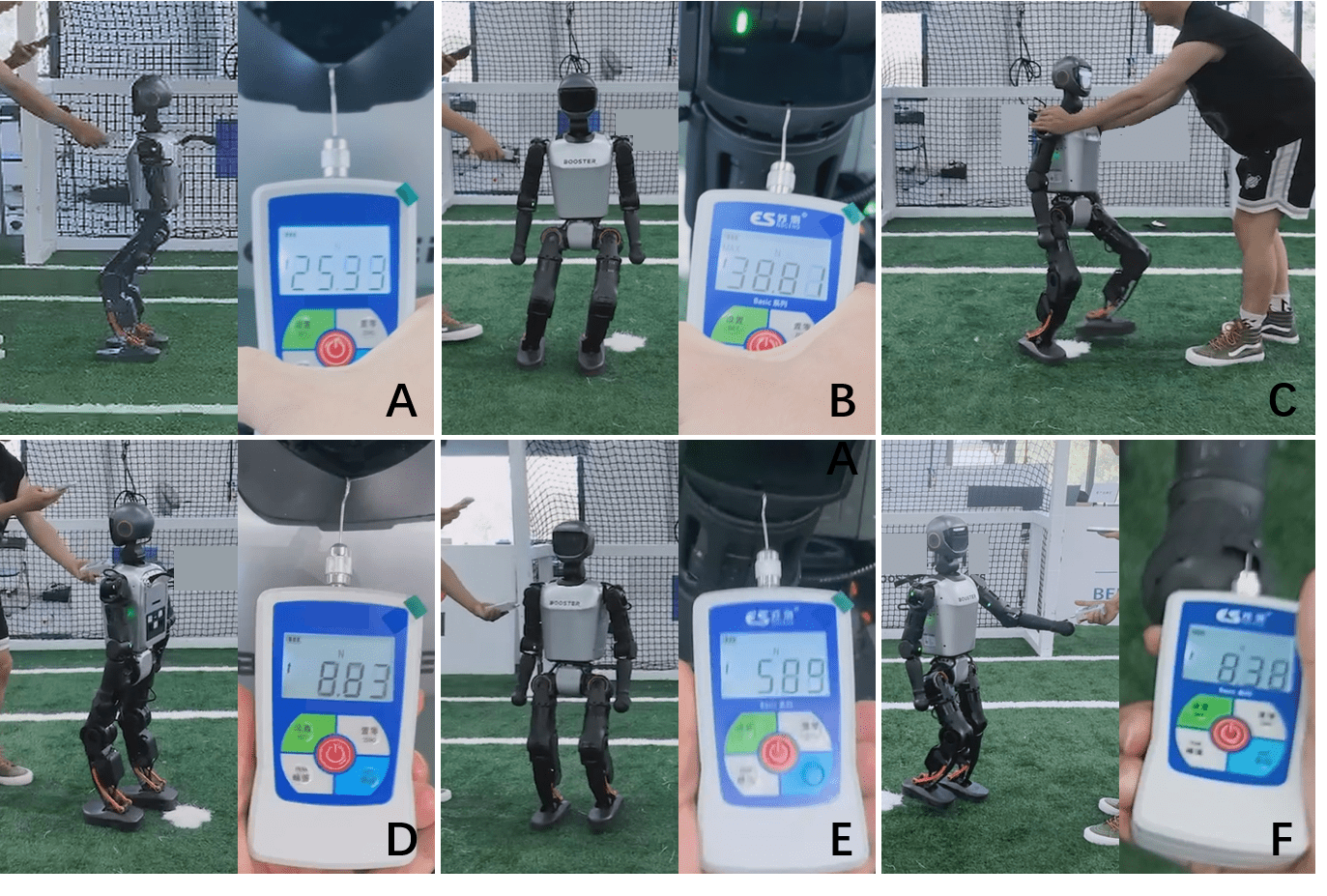}
    \caption{\textbf{Omnidirectional force-compliance walking:} (a)(b)(c) shows the baseline policy with force in x,y and yaw direction. (d)(e)(f) shows our policy with force in x,y and yaw direction. The screenshots represent typical reading of the dynamometer in each case.}
    \label{fig:reader}
    \vspace{-15pt}
\end{figure}

We further assess force compliance using quantitative real-world force measurements, addressing the ambiguity of manual human guidance. Using weights $w_c, w_f, w_r = [0.0, 2.0, 2.0]$, we pull the robot with minimal effort, measuring force with an external dynamometer \Cref{fig:reader}. The dynamometer is attached to the robot's neck, shoulder and hand with respect to forces in the x and y directions and torque around the z axis. Our MORL policy requires only about 10\,N to move the robot smoothly and efficiently, while the baseline demands over 25\,N and often exceeds the 30\,N measurement limit. Additionally, the baseline policy produces abrupt actions when resisting the applied torque, making it impossible to measure with dynamometer, while our policy still remains gentle and stable. These results indicate a lower force requirement and improved responsiveness under human-guided locomotion compared to the baseline.

\begin{figure}
    \vspace*{0.07in}
    \centering
    \includegraphics[width=1.0\linewidth]{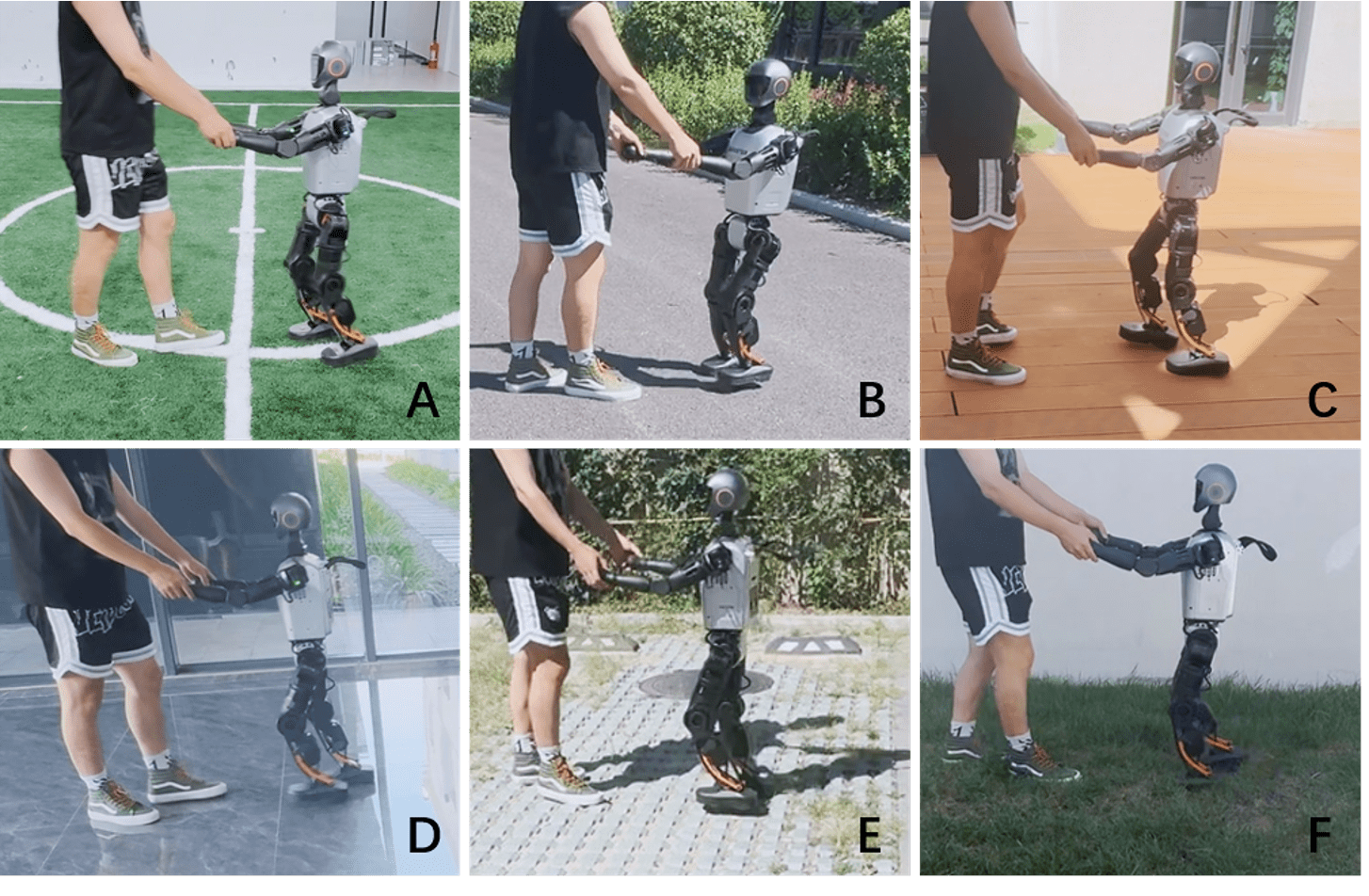}
    \caption{\textbf{Force-compliance walking in outdoor scenarios:} (a) soccer field, (b) rough ground, (c) wooden ground, (d) smooth surface, (e) raised ground, (f) soft grass. The robot is guided by the human operator pulling its arms. The shoulder joint are kept flexible with a small pre-set $k_p = 1$. No supportive force is exerted to the robot.}
    \label{fig:demo}
\end{figure}

The experiment in different outdoor environments is also conducted. In \Cref{fig:demo}, the robot is guided by hand-pulling and successfully traverses diverse terrains, including rough ground, a soccer field, raised surfaces, and others. The results qualitatively showcase the practicality and robustness of our policy. Across all scenarios, the robot exhibits smooth, stable, and compliant walking behaviors, responding reliably to subtle guiding forces applied by the human operator. In these outdoor demonstrations, the operator applies only horizontal guiding forces without providing any supportive or stabilizing assistance, ensuring that balance is maintained solely by the learned policy.

\subsection{Robustness to Instant Perturbation}
\begin{figure}
  \centering
  \begin{subfigure}[b]{0.49\linewidth}
    \centering
    \includegraphics[width=\linewidth]{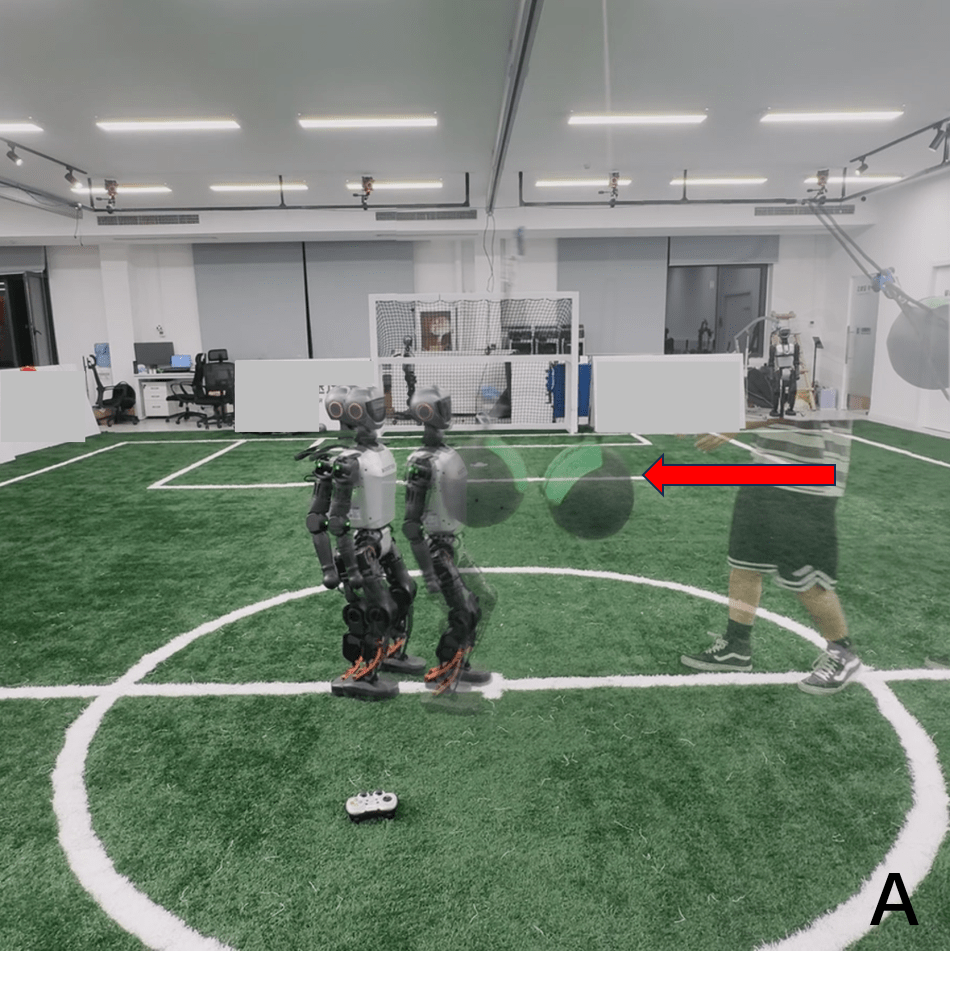}
  \end{subfigure}
  \hfill
  \begin{subfigure}[b]{0.49\linewidth}
    \centering
    \includegraphics[width=\linewidth]{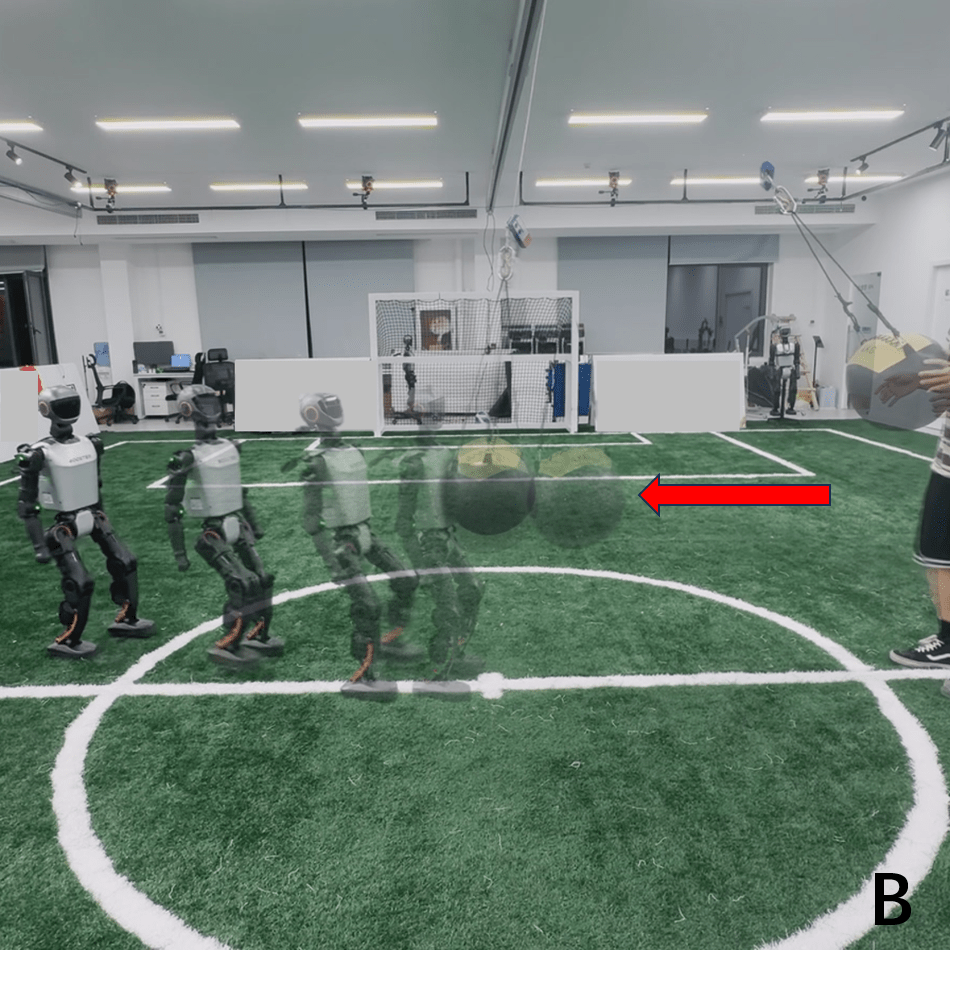}
  \end{subfigure}
  \caption{\textbf{Perturbation resistance under suspended-ball impacts}. The red arrow represents the direction of the strike.
    (a) Impact of a 2-kg ball, the robot maintains balance while maintaining position.  
    (b) Impact of a 4-kg ball, the robot steps back but remains stable.}
  \label{fig:ball_impact}
\end{figure}

To provide an additional indication of robustness on real hardware, we evaluate the system under sudden external impacts by swinging a suspended ball to strike the robot. As depicted in \Cref{fig:ball_impact}, a ball is hung from the ceiling and released manually to hit the robot. The preference weight is set to be $w_c, w_f, w_r = [1.0, 1.0, 2.0]$

The experiments show that the MORL policy can withstand unexpected impacts from balls weighing up to 5\,kg without falling. The robot adapts by taking backward steps and absorbing the force through compliant motion. These observations align with the disturbance tolerance seen in simulation, providing additional empirical support, though this test is not the primary focus of our work.

\section{Conclusion}
\label{sec:conclusion}

In this work, we introduced a preference-conditioned multi-objective RL framework for humanoid locomotion, which integrates command tracking and force compliance through a velocity–resistance model. The approach is simple and effective, requiring no hierarchical control or multi-stage training, and enables a single policy to interpolate smoothly between rigid tracking and compliant behaviors. An encoder–decoder architecture inferring force-related latent features further allows the policy to be deployable on real humanoid hardware without direct force sensing.

Simulation and hardware experiments demonstrate that our method could generate adaptive and force-compliant omnidirectional walking. Beyond demonstrating feasibility, this study highlights preference-conditioned MORL as a practical approach for interactive humanoid locomotion, with opportunities for future extensions to
richer tasks and higher-dimensional objectives.

\section*{Acknowledgment}
We would like to thank Booster Robotics for providing the T1 hardware, testing facilities, and technical support. We also thank Penghui Chen, Qingkai Li, and Xiaokang Sun for their assistance during the experiments.

\bibliographystyle{IEEEtran}
\bibliography{IEEEabrv,mybibfiles}

\end{document}